\documentclass{article}

\usepackage[final]{neurips_2025}
\usepackage{graphicx}
\usepackage{amsmath} 

\usepackage[utf8]{inputenc}
\usepackage[T1]{fontenc}
\usepackage{hyperref}
\usepackage{url}
\usepackage{booktabs}
\usepackage{amsmath, amsfonts}
\usepackage{nicefrac}
\usepackage{microtype}
\usepackage{xcolor}
\usepackage{array}
\usepackage{multirow}
\usepackage{longtable}
\usepackage{subcaption}
\usepackage{enumitem}
\usepackage{algorithm}
\usepackage{algpseudocode}
\newcommand{\REQUIRE}{\Require}

\newcommand{\STATE}{\State}
\newcommand{\IF}{\If}
\newcommand{\ELSE}{\Else}
\newcommand{\ENDIF}{\EndIf}
\newcommand{\RETURN}{\State \Return}
\newcommand{\COMMENT}[1]{\Comment{#1}}
\usepackage{makecell} 
\usepackage{subcaption}

\title{IPAD: Inverse Prompt for AI Detection - A Robust and Interpretable LLM-Generated Text Detector}

\author{
\textbf{Zheng Chen}\textsuperscript{1}\thanks{Equal contribution.}\,
\textbf{Yushi Feng}\textsuperscript{2}\footnotemark[1]\,
\textbf{Jisheng Dang}\textsuperscript{3}\,
\textbf{Yue Deng}\textsuperscript{1}\,
\textbf{Changyang He}\textsuperscript{4}\,\\
\textbf{Hongxi Pu}\textsuperscript{5}\,
\textbf{Haoxuan Li}\textsuperscript{6}\thanks{Corresponding authors.}\,
\textbf{Bo Li}\textsuperscript{1}\footnotemark[2]\\
\textsuperscript{1}Computer Science and Engineering, Hong Kong University of Science and Technology\\
\textsuperscript{2}School of Computing and Data Science, The University of Hong Kong\\
\textsuperscript{3}School of Information Science \& Engineering, Lanzhou University\\
\textsuperscript{4}Max Planck Institute for Security and Privacy\\
\textsuperscript{5}Computer Science, The University of Michigan\\
\textsuperscript{6}Center for Data Science, Peking University\\
\texttt{zchenin@connect.ust.hk, fengys@connect.hku.hk, dangjisheng@lzu.edu.cn, }\\
\texttt{ydengbi@connect.ust.hk, changyang.he@mpi-sp.org,}\\
\texttt{hongxi@umich.edu, hxli@stu.pku.edu.cn, bli@cse.ust.hk}\\
}

\begin{document}

\maketitle

\begin{abstract}
Large Language Models (LLMs) have attained human-level fluency in text generation, which complicates the distinguishing between human-written and LLM-generated texts. This increases the risk of misuse and highlights the need for reliable detectors. Yet, existing detectors exhibit poor robustness on out-of-distribution (OOD) data and attacked data, which is critical for real-world scenarios. Also, they struggle to provide interpretable evidence to support their decisions, thus undermining the reliability. In light of these challenges, we propose ~\textbf{IPAD (Inverse Prompt for AI Detection)}, a novel framework consisting of a ~\textbf{Prompt Inverter} that identifies predicted prompts that could have generated the input text, and two ~\textbf{Distinguishers} that examine the probability that the input texts align with the predicted prompts. Empirical evaluations demonstrate that IPAD outperforms the strongest baselines by 9.05\% (Average Recall) on in-distribution data, 12.93\% (AUROC) on out-of-distribution data, and 5.48\% (AUROC) on attacked data. IPAD also performs robustly on structured datasets. Furthermore, an interpretability assessment is conducted to illustrate that IPAD enhances the AI detection trustworthiness by allowing users to directly examine the decision-making evidence, which provides interpretable support for its state-of-the-art detection results.
\end{abstract}

\section{Introduction}\label{sec:Introduction}
Large Language Models (LLMs), characterized by their massive scale and extensive training data ~\citep{r1,r100,cheng2025empowering}, have achieved significant advances in natural language processing (NLP) ~\citep{r10,r11,r12}. However, with the advanced capabilities of LLMs, they are subject to frequent misused in various domains, including academic fraud, the creation of deceptive material, and the generation of fabricated information~\citep{r13,r14,r15,chen-etal-2025-llms-recognize}, which underscores the critical need to distinguish between human-written text (HWT) and LLM-generated text (LGT)~\citep{r14,r18,r19}.

However, due to their sophisticated functionality, LLMs pose significant challenges in the robustness of current AI detection systems~\citep{r12}. The existing detection systems, including commercial ones, frequently misclassify texts as HWT~\citep{r20,r21} and generate inconsistent results when analyzing the same text using different detectors~\citep{r22,r23}. Studies show false positive rates reaching up to 50\% and false negative rates as high as 100\% in different tools~\citep{r23} when dealing with out-of-distribution (OOD) datasets.

Another critical issue with the existing AI detection systems is their lack of verifiable evidence~\citep{r24, r200}, as these tools typically provide only simple outputs like \textit{"likely written by AI"} or percentage-based predictions~\citep{r23}. The lack of evidence prevents users from defending themselves against false accusations~\citep{r22} and hinders organizations from making judgments based solely on the detection results without convincing evidences~\citep{r23}. This problem is particularly troublesome not only because the low accuracy of such systems as mentioned before, but also due to the consequent inadequate response to LLM misuse, which can lead to significant societal harm~\citep{r25,r26,r27,r12}. These limitations highlight the pressing need for more reliable, explainable and robust detectors.

\begin{figure*}[t]
  \centering
  \includegraphics[width=\textwidth]{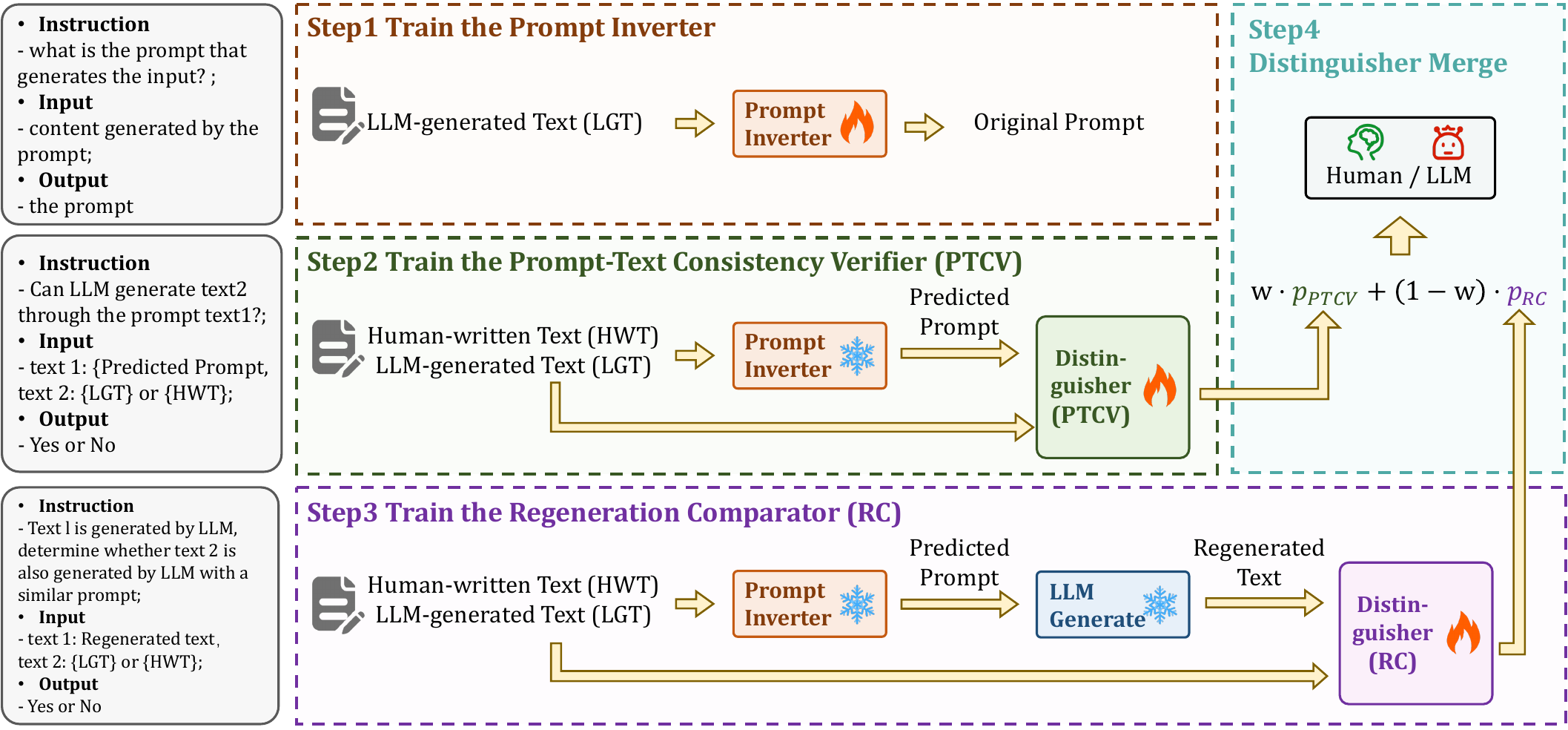}
  \caption{The overall workflow of our proposed IPAD framework}
  \label{fig:workflow}
\end{figure*}

In this paper, we propose \textbf{IPAD} (Inverse Prompt for AI Detection), a novel and interpretable framework for detecting AI-generated text. As illustrated in Figure~\ref{fig:workflow}, IPAD consists of two main components: a \textbf{Prompt Inverter}, which reconstructs the underlying prompts from input texts, and two \textbf{Distinguishers}—the \textbf{Prompt-Text Consistency Verifier (PTCV)}, which measures the alignment between the predicted prompt and input text, and the \textbf{Regeneration Comparator (RC)}, which compares the input with the corresponding regenerated text for consistency. By explicitly modeling the reasoning path from prompt inversion to final classification, IPAD introduces a paradigm shift in AI-generated content detection, significantly enhancing both detection robustness and user interpretability. 

Empirical results show that IPAD outperforms state-of-the-art baselines by 9.05\% in Average Recall on in-distribution datasets, 12.93\% in AUROC on out-of-distribution (OOD) datasets, and 5.48\% in AUROC under adversarial attacks. IPAD also generalizes well to structured data. A user study further reveals that IPAD improves trust and usability in detection tasks by presenting concrete decision evidence, including predicted prompts and regenerated texts. Code is available at \url{https://github.com/Bellafc/IPAD-Inver-Prompt-for-AI-Detection}.

Our contributions can be summarized as follows:
\begin{itemize}
  \item We introduce a novel fine-tuned inverse-prompt-based detection framework that integrates prompt reconstruction and dual consistency evaluation.
  \item We achieve superior detection performance on in-distribution, OOD,  adversarially attacked, and prompt-structured datasets.
  \item We demonstrate through an interpretability assessment that IPAD improves human trust and interpretability in AI text detection.
\end{itemize}

\vspace{-0.3cm}

\section{Methodology}\label{sec:Method}
\vspace{-0.3cm}

\subsection{Preliminaries}
\paragraph{Modules.} Our method comprises a \textbf{Prompt Inverter} ~\textbf{$f_{\text{inv}}$}, and two Distinguishers, namely the \textbf{Prompt-Text Consistency Verifier (PTCV)} \textbf{$f_{\text{PTCV}}$} and the \textbf{Regeneration Comparator (RC)} \textbf{$f_{\text{RC}}$}. Given an input text $T$, the task is to determine whether it is human-written (HWT) or generated by an LLM (LGT). We denote by $\mathcal{D}_{\text{PI}}$ the training set for $f_{\text{inv}}$, consisting of pairs $(T, P)$ where $T$ is an LLM-generated text and $P$ is its original prompt. The two distinguishers are trained using disjoint datasets: $\mathcal{D}_{\text{LGT}}$ contains LLM-generated samples and $\mathcal{D}_{\text{HWT}}$ contains human-written ones. All components are fine-tuned using Microsoft’s Phi3-medium-128k-instruct model\cite{DBLP:journals/corr/abs-2404-14219}.
\paragraph{Softmax-Based probability for Binary Classification in LLM.} 
\label{sec:softmax}
To estimate the fine-tuned model's binary classification probability (i.e., the probability of predicting ``yes'' or ``no''), we follow the logit-based estimation approach~\citep{yoshikawa-okazaki-2023-selective}. Given the model input $x$, and the output logits $z$, the model's probability assigned to $\hat{y}$ is computed through the softmax function $\sigma$ :
\[
\text{Confidence}_{\text{yes}} = P(\hat{y} = \text{``yes''} \mid x) = \sigma(z)_{\text{yes}}, \quad
\text{Confidence}_{\text{no}} = P(\hat{y} = \text{``no''} \mid x) = \sigma(z)_{\text{no}}
\]
Since the fine-tuned model will only output``yes'' or ``no'', we further calculate the probability for this binary classification as:
\[
\text{Probability}_{\text{yes}} = \frac{\text{Confidence}_{\text{yes}}}{\text{Confidence}_{\text{yes}} + \text{Confidence}_{\text{no}}}, \quad
\text{Probability}_{\text{no}} = 1 - \text{Probability}_{\text{yes}}
\]

\subsection{Workflow}
\vspace{-0.2cm}
Our framework follows a multi-stage fine-tuning pipeline with the following four steps, as illustrated in Figure~\ref{fig:workflow}. The details of the datasets for fine-tuning is illustrated in Appendix ~\ref{sec:dataset}.

\paragraph{Step 1: Training Prompt Inverter.} We first fine-tune a model $f_{\text{inv}}$ on dataset $\mathcal{D}_{\text{PI}}$, with the data structure shown in Figure ~\ref{fig:workflow}. For any input text $T$, $f_{\text{inv}}$ predicts the most likely prompt $P$ that could have generated it, i.e. $P = f_{\text{inv}}(T)$. The resulting Prompt Inverter is then frozen and reused in the following downstream steps.

\paragraph{Step 2: Training the Prompt-Text Consistency Verifier (PTCV).} Given the predicted prompt $P$ in step 1, and the input text $T \in \{\text{HWT}, \text{LGT}\}$, the verifier $f_{\text{PTCV}}$ is trained to predict whether the text $T$ could plausibly be generated by an LLM using the prompt $P$. The fine-tuning datasets $\mathcal{D}_{\text{LGT}}$  and $\mathcal{D}_{\text{HWT}}$  share the same structure, with output labels "yes" for $\mathcal{D}_{\text{LGT}}$  and "no" for $\mathcal{D}_{\text{HWT}}$, as shown in the Figure ~\ref{fig:workflow}.

After fine-tuning this module, we applied it to the validation set and computed the probability score $p_{\text{PTCV}} = f_{\text{PTCV}}(T, P)$, where the confidence value was estimated using the softmax-based method described in Section ~\ref{sec:softmax}.

\paragraph{Step 3: Training the Regeneration Comparator (RC).} With the same predicted prompt $P$ in step 1, we use an LLM to generate a regenerated text $T' \leftarrow LLM(P)$. By default, the LLM we use is \texttt{gpt-3.5-turbo}. Then, the comparator $f_{\text{RC}}$ is trained to assess whether $T$ and $T'$ can be generated by LLM with a similar prompt. This step uses the same dataset as in Step 2, but applies a different structural formatting, as shown in Figure ~\ref{fig:workflow}. 

After fine-tuning this module, we also applied it to the validation set and computed the probability score $p_{\text{RC}} = f_{\text{RC}}(T, P)$.

\paragraph{Step 4: Distinguisher Merge.} To determine the final classification, we combine the two probability scores, $p_{\text{PTCV}}$ and $p_{\text{RC}}$, obtained from Step 2 and Step 3 on the validation set. Specifically, we compute a weighted ensemble as $\hat{p} = w \cdot p_{\text{PTCV}} + (1 - w) \cdot p_{\text{RC}}$, and assign the prediction $\hat{Y} = \text{LGT}$ if $\hat{p} > \tau$, or $\hat{Y} = \text{HWT}$ otherwise. The weight $w \in [0,1]$ and the threshold $\tau \in [0,1]$ are treated as hyperparameters and selected via grid search on the validation set. The selected values were $w = 0.45$ and $\tau = 0.54$.

\paragraph{Inference.} We perform inference on unseen input texts $T$ by sequentially applying the trained modules. Given an input text $T$, we first use the prompt inverter $f_{\text{inv}}$ to recover the most plausible prompt $P$. The prompt is then used to regenerate a candidate text $T'$ via the an LLM. Next, we compute two probability scores: $p_{\text{PTCV}}$, indicating whether $T$ is consistent with $P$, and $p_{\text{RC}}$, assessing the likelihood that $T$ and $T'$ originate from the same prompt. These scores are fused into a final decision score $\hat{p}$ using the gird-searched weight $w$, and the predicted label is determined by comparing $\hat{p}$ against the threshold $\tau$. The complete inference pipeline is summarized in Algorithm~\ref{alg:ipad}.

\begin{minipage}[t]{0.48\textwidth}
\vspace{-5mm}
\begin{algorithm}[H]
\caption{\textsc{IPAD} Detection Procedure}
\label{alg:ipad}
\begin{algorithmic}[1]
\REQUIRE Input text $T$; trained modules $\mathbf{f}_{\text{inv}}, \mathbf{f}_{\text{PTCV}}, \mathbf{f}_{\text{RC}}$; LLM $\mathbf{f}_{\text{LLM}}$; fusion weight $w \in [0, 1]$; threshold $\tau \in [0, 1]$
\STATE $P \leftarrow \mathbf{f}_{\text{inv}}(T)$ \COMMENT{Inverse-prompt prediction}
\STATE $T' \leftarrow \mathbf{f}_{\text{LLM}}(P)$ \COMMENT{Regenerate text using $P$}
\STATE $\boldsymbol{z}^{\text{PTCV}} \leftarrow \mathbf{f}_{\text{PTCV}}(P, T)$
\STATE $p_{\text{PTCV}} \leftarrow \frac{\sigma(\boldsymbol{z}^{\text{PTCV}}_{\text{yes}})}{\sigma(\boldsymbol{z}^{\text{PTCV}}_{\text{yes}}) + \sigma(\boldsymbol{z}^{\text{PTCV}}_{\text{no}})}$
\STATE $\boldsymbol{z}^{\text{RC}} \leftarrow \mathbf{f}_{\text{RC}}(T', T)$
\STATE $p_{\text{RC}} \leftarrow \frac{\sigma(\boldsymbol{z}^{\text{RC}}_{\text{yes}})}{\sigma(\boldsymbol{z}^{\text{RC}}_{\text{yes}}) + \sigma(\boldsymbol{z}^{\text{RC}}_{\text{no}})}$
\STATE $\hat{p} \leftarrow w \cdot p_{\text{PTCV}} + (1 - w) \cdot p_{\text{RC}}$
\IF{$\hat{p} > \tau$}
    \STATE $\hat{Y} \leftarrow \mathrm{LGT}$
\ELSE
    \STATE $\hat{Y} \leftarrow \mathrm{HWT}$
\ENDIF
\STATE $\mathcal{E} \leftarrow (P,\; p_{\text{PTCV}},\; p_{\text{RC}},\; \hat{p})$
\RETURN $(\hat{Y},\; \mathcal{E})$
\end{algorithmic}
\end{algorithm}
\end{minipage}%
\hfill
\begin{minipage}[t]{0.5\textwidth}
\subsection{Computational Complexity and Deployment Considerations}

The inference procedure of the IPAD framework consists of three calls through a light-weight open-sourced LLM \texttt{phi-3-medium-128k-instruct}. \texttt{Phi-3} is a decoder-only Transformer, whithin which, the self-attention complexity per layer is $\mathcal{O}(n^2 \cdot d)$, where $n$ is the sequence length and $d$ is the hidden dimension~\citep{vaswani2017attention}. The additional api call to \texttt{gpt-3.5-turbo} for re-generating texts introduces fixed latency but no local computation cost. Therefore, the overall computational cost is bounded by \(\mathcal{O}(3 \cdot L \cdot n^2 \cdot d + \texttt{OpenAI}_{\text{api}})\), where \(L = 32\) is the number of layers in \texttt{phi-3}~\citep{DBLP:journals/corr/abs-2404-14219}, which is relatively small. All three \texttt{phi-3} calls can be deployed in an Nvidia V100 GPU as the minimum requirement. This demonstrates that IPAD is not computationally expensive and can be deployed with relatively modest hardware requirements.
\end{minipage}

\vspace{1em}

\subsection{Training}
The supervised fine-tuning ~\citep{r70} process is performed on a Microsoft's open model, \texttt{phi3-medium-128k-instruct}, and we use low-rank adaptation (LoRA) method ~\citep{r69} on the \texttt{LLaMA-Factory} framework~\citep{r68}. We train it using six A800 GPUs for 20 hours for \textbf{Prompt Inverter}, 7 hours for \textbf{PTCV}, and 9 hours for \textbf{RC}. 
\section{Experiments}\label{sec:Evaluation}
We investigate the following questions through our experiments:  

\begin{itemize}[itemsep=1pt, topsep=1pt]
  \item Assess the robustness of IPAD, which includes using various LLMs as generators, comparing IPAD with other detectors, and evaluating on out-of-distribution (OOD), attacked datasets, and prompt-structured datasets.  
  \item Independently analyze the necessity and effectiveness of the \textbf{Prompt Inverter}, the \textbf{PTCV}, and the \textbf{RC}.  
  \item Explore the user-friendliness of IPAD through an interpretability assessment.
\end{itemize}

\subsection{Robustness of IPAD}
\subsubsection{Evaluation Baselines and Metrics}
The in-distribution experiments refer to the testing results presented in ~\citep{r3}, where the data aligns with the training data used for the IPAD, thereby serving as our baseline. This baseline assesses how the RoBERTa classifiers (base and large)~\citep{park2021klue}, the HC3 detector~\citep{hc3}, and the OUTFOX detector~\citep{r3} perform on standard data as well as under DIPPER~\citep{alkanhel2023dipper} and OUTFOX attacks.

The OOD experiments refer to the DetectRL baseline ~\citep{r58}, which is a comprehensive benchmark, which includes four datasets: (1) academic abstracts from the arXiv Archive (covering the years 2002 to 2017)
, (2) news articles from the XSum dataset ~\citep{r59}, (3) creative stories from Writing Prompts ~\citep{r60}, and (4) social reviews from Yelp Reviews~\citep{r61}. It also employs three attack methods to simulate complex real-world detection scenarios, which include (1) the prompt attacks, (2) paraphrase attacks, and (3) perturbation attacks~\citep{r58}. DetectRL evaluates three classifiers on the OOD dataset: DetectLLM (LRR)~\citep{su2023detectllm}, Fast-DetectGPT~\citep{bao2023fast},  RoBERTa Classifier (Base). We included two more strong classifiers in our evaluation DetectLLM (NPR)~\citep{su2023detectllm} and Binoculars ~\citep{hans2024spotting}. All the testing sets have 1,000 samples in our experiments.

We further evaluate its performance on OOD datasets with \textbf{structured prompts}. The LongWriter dataset~\citep{bai2025longwriter}, featuring an average prompt length of 1,501 tokens, reflects IPAD's capability to handle long-form prompts. The Code-Feedback\citep{r101} and Math datasets~\citep{hendrycks2021measuring} contain highly structured prompts, in contrast to typical essay-like writing. We compare IPAD with baseline detectors from DetectRL to assess its relative performance under these challenging conditions.

The ~\textbf{Area Under Receiver Operating Characteristic curve (AUROC)} is widely used for assessing detection method ~\citep{r55}. Since our models predict binary labels, we follow the ~\textit{Wilcoxon-Mann-Whitney} statistic~\citep{r56}, and the formula is shown in Appendix ~\ref{sec:AUROC formula}. 
The ~\textbf{AvgRec} is the average of ~\textbf{HumanRec} and ~\textbf{MachineRec}, which refers to the recall of the Human-written texts and the LLM-generated texts~\citep{r57}. 

\subsubsection{Robustness across different LLMs}

As shown in Table~\ref{tab:ipad_results}, IPAD achieves consistently strong performance across all combinations of original generators and re-generators, which shows its robustness to diverse LLM as generators. The best results are generally observed when the original generator and the re-generator are the same, while the \texttt{gpt-3.5-turbo} serves as an effective universal re-generator: it performs well even when the original generator differs. In real-world applications where the identity of the original generator is unknown, using \texttt{gpt-3.5-turbo} as a fixed re-generator provides a practical and reliable solution.

\begin{table}[t]
\centering
\caption{Detection Accuracy (HumanRec, MachineRec, AvgRec, and AUROC \%) of IPAD across Various LLMs on In-Distribution Data}
\label{tab:ipad_results}
\resizebox{0.9\textwidth}{!}{

\begin{tabular}{ll|cccc}
\toprule
\textbf{Original Generator} & \textbf{Re-Generator} & \textbf{HumanRec} & \textbf{MachineRec} & \textbf{AvgRec} & \textbf{AUROC} \\
\midrule
\texttt{gpt-3.5-turbo} & \texttt{gpt-3.5-turbo}& 98.50 & 100 & 99.25 &  100\\
\hline
\texttt{gpt-4} & \texttt{gpt-4} & 98.70 & 100 & 99.35 & 100 \\
 & \texttt{gpt-3.5-turbo} & 96.10 & 100 & 98.05 &  99.96\\
 \hline
\texttt{Qwen-turbo} & \texttt{Qwen-turbo} & 98.60 & 99.80 & 99.20&99.96\\
 & \texttt{gpt-3.5-turbo} & 98.40 & 99.50 & 98.95&99.86\\
 \hline
\texttt{LLaMA-3-70B} & \texttt{LLaMA-3-70B} & 98.70 & 100 & 99.35 & 100 \\
 & \texttt{gpt-3.5-turbo} & 98.60 & 100 & 99.30 &  100\\
\bottomrule
\end{tabular}}

\end{table}

\subsubsection{Comparison of IPAD with other detectors in and out of distribution}
\begin{figure}[t]
  \centering
  \includegraphics[width=\textwidth]{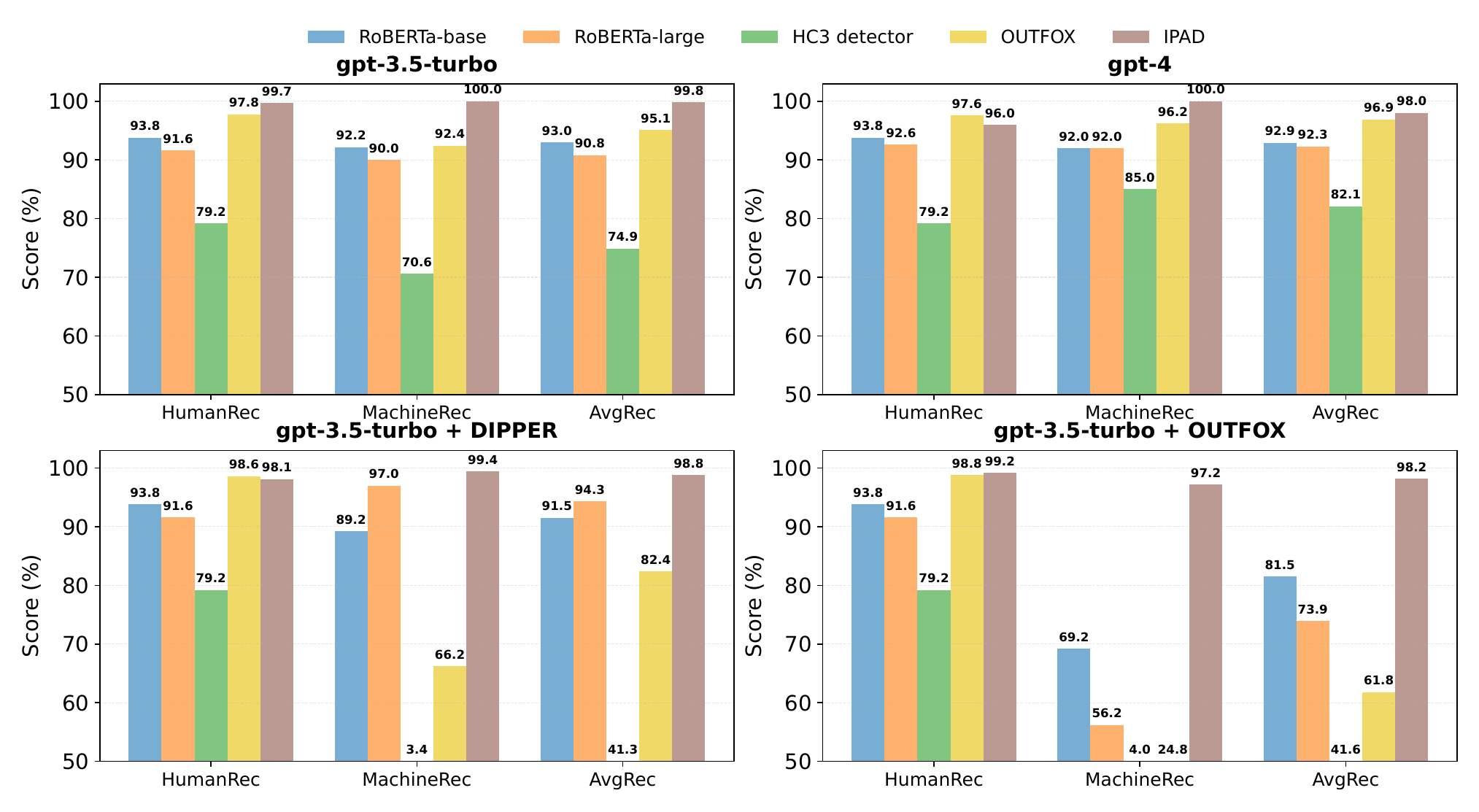}
  \caption{The In-distribution data performance of IPAD and the baseline detectors. Since ~\cite{r3} only presents the AvgRec data for the baselines, we also calculate AvgRec data for IPAD to compare.}
  \label{fig:indis}
\end{figure}

\paragraph{In Distribution. } For the in-distribution data, as shown in Figure~\ref{fig:indis}, the baseline detectors like RoBERTa, HC3, and OUTFOX perform well on standard data but degrade significantly under DIPPER and OUTFOX attacks. In contrast, IPAD maintains high accuracy across all scenarios, which surpasses the strongest baseline \textbf{9.05\%} in AvgRec.

\paragraph{Out of Distribution. }
 
Table~\ref{tab:baseline} reports detection accuracy across four benchmark datasets, which shows that IPAD significantly outperforms prior baselines. Table~\ref{tab:attack-results} further evaluates robustness under three attack types, where IPAD again demonstrates superior resilience. Compared to the strongest baseline, IPAD achieves a \textbf{12.93\%} relative improvement on standard datasets in AUROC and a \textbf{5.48\%} improvement on attack datasets.

\begin{table}[h]
\centering
\caption{Detection Accuracy (AUROC \%) on four diverse OOD datasets}
\label{tab:baseline}
\resizebox{0.8\textwidth}{!}{
\begin{tabular}{lcccc|c}
\toprule
\textbf{Method} & \textbf{Arxiv} & \textbf{XSum} & \textbf{Writing} & \textbf{Review} & \textbf{Average} \\
\midrule
DetectLLM (LRR) & 48.17 & 48.41 & 58.70 & 58.21 & 53.37 \\
DetectLLM (NPR) & 53.85 & 34.59 & 54.96 & 50.09 & 48.37 \\
Binoculars & 84.03 & 77.39 & 94.38 & 90.00 & 86.45 \\
Fast-DetectGPT & 42.00 & 45.72 & 51.13 & 54.55 & 48.35 \\
Rob-Base & 81.06 & 76.81 & 86.29 & 87.84 & 83.00 \\
IPAD Merge & \textbf{100} & \textbf{99.85} & \textbf{99.40} & \textbf{98.25} & \textbf{99.38} \\
\bottomrule
\end{tabular}}

\end{table}

\begin{table}[ht]
  \centering
    \caption{Detection Accuracy (AUROC \%) on three attacked OOD datasets}
  \label{tab:attack-results}
  \resizebox{0.9\textwidth}{!}{
  \begin{tabular}{lccc|c}
    \toprule
    \textbf{Method} & \textbf{Prompt Attack} & \textbf{Paraphrase Attack} & \textbf{Perturbation Attack} & \textbf{Average} \\
    \midrule
    DetectLLM (LRR) & 54.97 & 49.23 & 53.62 & 52.61 \\
    DetectLLM (NPR) & 77.15 & 56.94 & 6.78 & 46.96 \\
    Binoculars      & 93.45 & 88.34 & 76.89 & 86.23 \\
    Fast-DetectGPT  & 43.89 & 41.15 & 44.38 & 43.14 \\
    Rob-Base        & 92.81 & 90.02 & 92.12 & 91.65 \\
    IPAD      & \textbf{97.30} & \textbf{96.00} & \textbf{98.10} & \textbf{97.13} \\
    \bottomrule
  \end{tabular}}

\end{table}

\vspace{-0.3cm}
\paragraph{Structured Prompts.} The results are shown in Table ~\ref{tab:structure}, while these datasets lack HWT references and are thus only evaluated using MachineRec, the strong scores suggest that IPAD maintains robustness even on structured diverse inputs, with an improvement of 9.87\% against the strongest baseline in MachineRec.

\begin{table}[ht]
  \centering
    \caption{Detection Accuracy (MachineRec \%) on three structured OOD datasets}
  \label{tab:structure}
  \resizebox{0.8\textwidth}{!}{
  \begin{tabular}{lccc|c}
    \toprule
    \textbf{Method} & \textbf{LongWriter} & \textbf{Code-Feedback} & \textbf{Math} & \textbf{Average} \\
    \midrule
    DetectLLM (LRR) & 32.1 & 29.0 & 30.2 & 30.43 \\
    DetectLLM (NPR) & 41.2 & 45.9 & 56.0 & 47.7 \\
    Binoculars      & 82.1 & 84.6 & 89.4 & 85.4 \\
    Fast-DetectGPT  & 12.0 & 11.1 & 15.1 & 12.7 \\
    Rob-Base        & 81.5 & 89.2 & 82.1 & 84.3 \\
    IPAD      & \textbf{97.5} & \textbf{92.7} & \textbf{95.6} & \textbf{95.27} \\
    \bottomrule
  \end{tabular}}

\end{table}

\subsection{Necessity and Effectiveness of the \textbf{Prompt Inverter}, \textbf{PTCV}, and \textbf{RC}}

\subsubsection{Necessity}
To prove that it is necessary to fine-tune on IPAD with IPAD with ~\textbf{PTCV} and ~\textbf{RC}, we conducted ablation study to use the same finetune method on only ~\textbf{input texts} and only ~\textbf{predicted prompts}, with the finetune data format shown in Appendix ~\ref{sec:iopo}. We only experimented on \textbf{Prompt Inverter + PTCV} and \textbf{Prompt Inverter + RC} to compare with the three-moduled IPAD.

\begin{figure}[t]
  \centering
  \includegraphics[width=\textwidth]{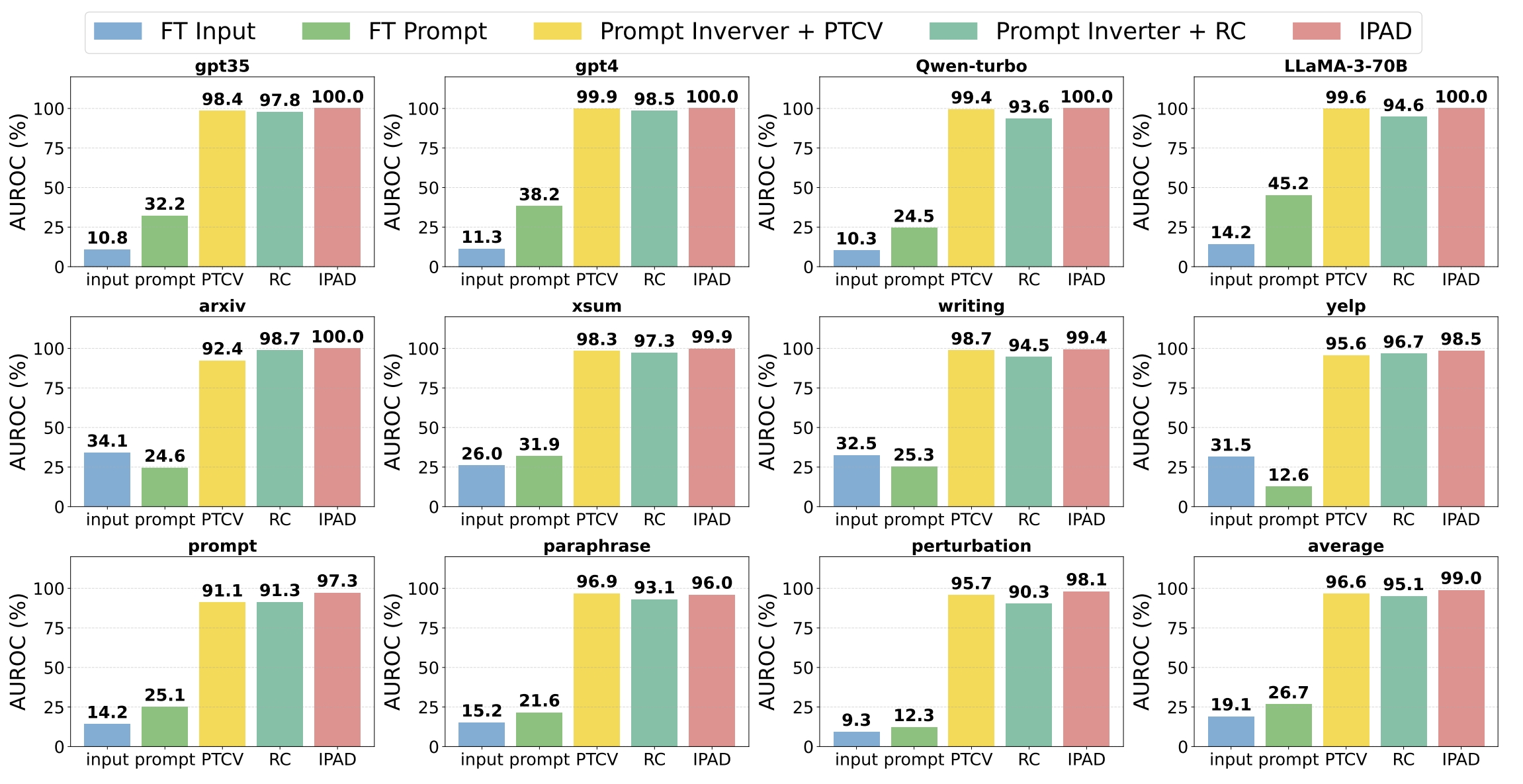}
  \caption{Ablation study. Evaluating \textbf{Fine-tune only on Input}, \textbf{Fine-tune only on Prompt}, \textbf{Prompt Inverter + PTCV}, \textbf{Prompt Inverter + RC}, and \textbf{IPAD} on In-distribution datasets, standard OOD datasets, and attacked OOD datasets.}
  \label{fig:ablation}
\end{figure}

Based on the ablation study results as shown in Figure ~\ref{fig:ablation}, fine-tuning only on input texts or only on predicted prompts performs poorly across all datasets in AUROC scores. While using \textbf{Prompt Inverter + PTCV} or \textbf{Prompt Inverter + RC} individually significantly improves performance, neither approach consistently excels across both HWT-style and LGT-style generations. In contrast, the full IPAD framework achieves consistently high performance across all settings, which demonstrates the necessity of the \textbf{Prompt Inverter}, \textbf{PTCV}, and \textbf{RC} modules.
\vspace{-0.3cm}

\begin{table}[t]
\centering
\caption{Comparison of prompt inverters on the similarities of the original prompts and the predicted prompts on LGT and HWT.}
\label{tab:1}
\resizebox{0.8\textwidth}{!}{
\begin{tabular}{l|lll|lll}
\hline
\multirow{2}{*}{\textbf{Metric}} & \multicolumn{3}{c|}{\textbf{LGT}} & \multicolumn{3}{c}{\textbf{HWT}} \\
\cline{2-7}
& DPIC & PE & IPAD & DPIC & PE & IPAD \\
\hline
Bart-large-cnn & -2.12 & -2.23 & \textbf{-1.84} & -2.47 & -2.39 & \textbf{-2.22} \\
Sentence-Bert  & 0.46  & 0.58  & \textbf{0.69}  & 0.42  & 0.53  & \textbf{0.57} \\
BLEU           & 5.61E-05 & 3.21E-04 & \textbf{0.24} & 8.75E-06 & 2.56E-08 & \textbf{0.13} \\
ROUGE-1        & 0.04  & 0.25  & \textbf{0.51}  & 0.06  & 0.13  & \textbf{0.39} \\
\hline
\end{tabular}
}

\end{table}

\begin{table}[t]
\centering
\caption{Comparison of distinguishers on HumanRec, MachineRec, and AvgRec (\%).}
\label{tab:2}
\resizebox{0.8\textwidth}{!}{
\begin{tabular}{l|lll}
\hline
\textbf{Distinguish Method} & \textbf{HumanRec} & \textbf{MachineRec} & \textbf{AvgRec} \\
\hline
Sentence-Bert (Thr. 0.67) & 61.20 & 95.20 & 78.20 \\
Bart-large-cnn (Thr. -2.52) & 42.60 & 97.20 & 69.90 \\
Prompt to ChatGPT & 33.20 & 64.50 & 48.85 \\
IPAD & \textbf{98.50} & \textbf{100.00} & \textbf{99.25} \\
\hline
\end{tabular}
}

\end{table}

\subsubsection{Effectiveness}

\paragraph{Prompt Inverter.} We use DPIC~\citep{r62} and PE~\citep{r65} as baseline methods for prompt extraction. DPIC employs a zero-shot approach using the prompt states in Appendix ~\ref{sec:DPIC prompt}, while PE uses adversarial attacks to recover system prompts. In our evaluation, we tested 1000 LGT and 1000 HWT samples. We use only in-distribution data for testing since only these datasets include original prompts. The metrics are all tested on comparing the similarity of the original prompts and the predicted prompts. The results shown in Table ~\ref{tab:1} illustrate that IPAD consistently outperforms both DPIC and PE across all four metrics (BartScore~\citep{r64}, Sentence-Bert Cosine Similarity~\citep{r63}, BLEU~\citep{r66}, and ROUGE-1~\citep{r67}), which highlight the effectiveness of the IPAD ~\textbf{Prompt Inverter}.

\vspace{-0.3cm}

\paragraph{PTCV and RC. }We conducted a comparison study using the frozen Prompt Inverter but different distinguishing methods. The first and second methods employed \texttt{Sentence-Bert}~\citep{r63} and \texttt{Bart-large-cnn}~\citep{r64} to compute the similarity score between the input texts and the regenerated texts. We selected thresholds that maximized AvgRec, which were 0.67 for Sentence-Bert and -2.52 for Bart-large-cnn. The classification rule is that the texts with scores greater than the threshold will be classified as LGT, while the texts with scores less than or equal to the threshold will be classified as HWT. The third method is to directly prompt ChatGPT in Appendix ~\ref{sec:iopo}, which mimic the fine-tuning process of \textbf{PTCV} and \textbf{RC}. The final results shown in Table ~\ref{tab:2} demonstrate that the other distinguishing methods performed worse than IPAD, highlighting the superior effectiveness of the IPAD ~\textbf{Distinguishers}.

\vspace{-0.3cm}

\paragraph{Compare with DPIC.} DPIC first uses a zero-shot prompt inverter to generate prompts, then applies a Siamese encoder and classifier to measure similarity between the embeddings of the original and regenerated texts. However, the classifier's reliance on embedding similarity is ambiguous, as similar texts may stem from different prompts. IPAD addresses this by fine-tuning directly on raw texts and reformulating the task as a logical reasoning problem as shown in the instructions of \textbf{PTCV} and \textbf{RC}. Our trained \textbf{Prompt Inverter} outperforms DPIC’s generic zero-shot method as shown in Table~\ref{tab:1}, and IPAD also achieves better performance than DPIC overall, as results shown in Appendix ~\ref{sec:compare dpic}.

\subsection{Interpretability Assessment of IPAD}

To assess the explainability improvement of IPAD, we designed an interpretability assessment with ten participants evaluating one HWT and one LGT article. We used IPAD version 2 due to its superior OOD performance and attack resistance. Participants compared three online detection platforms (i.e., Scribbr, QuillBot, GPTZero) with IPAD's process (which displayed input texts, predicted prompts, regenerated texts, and final judgments). After evaluation, participants rated IPAD on four key explainability dimensions. Transparency received strong ratings (40\%:5, 60\%:4), with participants appreciating the visibility of intermediate processes. Trust scores were more varied (10\%:3, 70\%:4, 20\%:5), but IPAD was generally considered more convincing than single-score detectors. Satisfaction was mixed (30\%:3, 30\%:4, 40\%:5), with participants acknowledging better detection but raising concerns about energy efficiency since IPAD runs three LLMs. Debugging received unanimous 5s, as participants could easily analyze the predicted prompt and regenerated text to verify the decision-making process. If needed, users could refine the generated content by adjusting instructions, such as specifying a word count, making IPAD a more effective and user-friendly tool compared to black-box detectors.

\section{Related Work}\label{sec:Related Work}
\subsection{AI detectors Methods and challenges}
Recent studies have explored diverse strategies for detecting AI-generated text. \textbf{Watermarking} embeds identifiable patterns during training~\cite{r28,r27} or inference~\cite{r29}, but requires model access and is vulnerable to erasure attacks~\cite{r30}. \textbf{Statistics-based methods} treat output distributions as detection signals. DetectGPT~\cite{r35} and Fast-DetectGPT~\cite{bao2023fast} locate LGT in regions of negative curvature of log-probability; Lastde~\cite{xu2025trainingfree} and Glimpse~\cite{bao2025glimpse} exploit token-probability dynamics and partial-distribution prediction. Other statistical approaches rely on n-gram divergence or revision similarity~\cite{r32,r31,r38,r37,r36}, though robustness remains limited~\cite{r12}. \textbf{Regeneration-based methods} compare model rewrites with originals: RAIDAR~\cite{r37} and MAGRET~\cite{huang-etal-2025-magret} observe stronger edits on human text; DNA-GPT~\cite{r36} and TOCSIN~\cite{r11} measure continuation or deletion-based differences. \textbf{Neural approaches} fine-tune large encoders (e.g., RoBERTa~\cite{r42}, BERT~\cite{r44}, XLNet~\cite{r45}) with adversarial or contrastive objectives~\cite{r14,r46}, while \textbf{human-in-the-loop methods} provides complementary semantic judgment and explainability~\cite{r22,r50,r49}.

\subsection{Prompt Inverter techniques and applications}
Prompt extraction techniques aim to reverse-engineer the prompts that generate specific outputs from LLMs. Approaches include black-box methods like output2prompt~\cite{r4}, which extracts prompts based on model outputs without access to internal data, and logit-based methods like logit2prompt~\cite{r51}, which rely on next-token probabilities but are constrained by access to logits. Adversarial methods can bypass some defenses but are model-specific and fragile~\cite{r52}. Despite the success of some zero-shot LLM-inversion based methods~\cite{r53, r62}, they are mostly naive usage of prompting LLMs, which makes them poor in prompt extraction accuracy and robustness.
\section{Conclusion}\label{sec:Conclusion}
\vspace{-0.3cm}
This paper introduces ~\textbf{IPAD (Inverse Prompt for AI Detection)}, a framework consisting of a ~\textbf{Prompt Inverter} that identifies predicted prompts that could have generated the input text, and two ~\textbf{Distinguishers} that examines how well the input texts align with the predicted prompts. One is the ~\textit{Prompt-Text Consistency Verifier (PTCV)} which evaluates direct alignment between predicted prompts and input text, and the other is ~\textit{Regeneration Comparator (RC)} that examines content similarity by comparing input texts with the corresponding regenerated texts. Empirical evaluations demonstrate that IPAD outperforms the strongest baselines by 9.05\% (Average Recall) on in-distribution data, 12.93\% (AUROC) on out-of-distribution (OOD) data, and 5.48\% (AUROC) on attacked data. IPAD also performs robustly on structured datasets. While the local alignment in \text{RC} approach provides explicit interpretability, it is more sensitive to adversarial attacks. In contrast, the global distribution in \textbf{PTCV} matching approach implicitly learns generative LLM's distributional properties, which offers more robustness while maintaining explainability. The combination of the two modules suggests that combining self-consistency checks of generative models with multi-step reasoning for evidential explainability holds promise for future AI detection systems in real-world scenarios. An interpretability assessment reveals that IPAD enhances trust and transparency by allowing users to examine decision-making evidence. Overall, IPAD establishes a new paradigm for more robust, reliable, and interpretable AI detection systems to combat the misuse of LLMs.

While IPAD demonstrates SOTA performance, two limitations warrant discussion:
(1) The \textbf{Prompt Inverter} may not fully reconstruct prompts containing explicit in-context learning examples, as it prioritizes semantic alignment over precise syntactic replication.
(2) While IPAD achieves strong performance across diverse datasets, it relies on LLMs, making it more computationally expensive compared to lightweight detectors such as RoBERTa or HC3. However, compared other detectors compared with LLMs, such as DPIC, IPAD is more lightweight since it calls the open-sources light-weight \texttt{Phi-3} model.

\section*{Acknowledgments}
This work was supported in part by NSFC grants (62432008 and 623B2002), in part by RGC RIF grant R6021-20, in part by RGC TRS grant T43-513/23N-2, in part by RGC CRF grants (C7004-22G, C1029-22G and C6015-23G), in part by NSFC/RGC grant CRS\_HKUST601/24, and in part by RGC GRF grants (16207922, 16207423 and 16203824).

\bibliographystyle{unsrtnat}
\bibliography{custom}


\appendix
\section{Fine-tune Dataset}
\label{sec:dataset}
\paragraph{Prompt Inverter Dataset.} We use the following four datasets, with the first three datasets enhance the model's generalization to recover the prompts, while the last dataset improves performance on essay-related tasks.
\begin{itemize}[]
\item \textbf{Instructions-2M}~\cite{r5}, a collection of 2 million user prompts and system prompts, from which we used 30,000 prompts.
\item \textbf{ShareGPT}~\cite{r6}, an open platform where users share ChatGPT prompts and responses, from which we used 500 samples.
\item \textbf{Unnatural Instructions}~\cite{r6}, a dataset of creative instructions generated by OpenAI’s models, from which we used 500 samples.
\item \textbf{OUTFOX dataset}~\cite{r3}, which contains 15,400 essay problem statements, student-written essays, and LLM-generated essays.
\end{itemize}
The first three datasets aims to enhance the general querying capability of the \textbf{Prompt Inverter}, and are all released under the MIT license. All the samples we used are the same to the samples randomly selected in ~\cite{r4}. The last dataset aims to enhance the familiarity of the \textbf{Prompt Inverter} with the data of the essay to detect the LLM-generated essays, and are created and examined by ~\citet{r3}, We specifically used the LLM-generated essays and problem statements for this supervised fine-tuning (SFT). There are 45,400 training pairs in total.

Given that essay data are diverse, we utilize only the OUTFOX dataset~\cite{r3}. To adapt this dataset for training our \textbf{Distinguisher}, we enhance it to align with the model's requirements. The original dataset consists of 14,400 training triplets of essay problem statements, student-written essays, and LLM-generated essays. To further process the data, we apply the \textbf{Prompt Inverter} to both student-written and LLM-generated essays, generating corresponding \textit{Predicted Prompts}. These \textit{Predicted Prompts} are then used to regenerate texts via \textbf{ChatGPT}, i.e. \textbf{gpt-3.5-turbo}. Following this procedure, we construct a total of 28,800 training samples, with an equal distribution of positive and negative examples (14,400 each).

The final dataset is structured as follows:  
\begin{table}[h]
\centering
\caption{Instruction, input/output structure, and inference outputs of each fine-tuned module. $T$ is the input text, $P$ the predicted prompt, and $T'$ the regenerated text.}
\label{tab:io_all_modules_transposed}
{\small
\begin{tabular}{p{1.5cm}|p{3.5cm}p{3.5cm}p{3.5cm}}
\toprule
\textbf{Field} & \textbf{Prompt Inverter} & \textbf{PTCV} & \textbf{RC} \\
\midrule
\textbf{Instruction} & \texttt{"What is the prompt $P$ that generates the Input Text $T$?"} & \texttt{"Can LLM generate the input text $T$ through the prompt $P$?"} & \texttt{"$T'$ is generated by LLM, determine whether $T$ is also generated by LLM with a similar prompt."} \\
\textbf{Input}       & $T$ & $(P, T)$ & $(T', T)$ \\
\textbf{Output}      & $P$ & \texttt{"yes"/"no"} & \texttt{"yes"/"no"} \\
\midrule
\textbf{Output in Inference} & $P$ & $p_{\text{PTCV}}$ & $p_{\text{RC}}$ \\
\bottomrule
\end{tabular}}
\vspace{0.5em}

\end{table}

\section{AUROC formula}
\label{sec:AUROC formula}
Since our model predicts binary labels, we follow the ~\textit{Wilcoxon-Mann-Whitney} statistic~\cite{r56} to calculate the Area Under Receiver Operating Characteristtic curve (AUROC):

\[
\text{AUC}(f) = \frac{\sum_{t_0 \in \mathcal{D}^0} \sum_{t_1 \in \mathcal{D}^1} \mathbf{1}[f(t_0) < f(t_1)]}{|\mathcal{D}^0| \cdot |\mathcal{D}^1|}
\]

where \( \mathbf{1}[f(t_0) < f(t_1)] \) denotes an indicator function which returns 1 if \( f(t_0) < f(t_1) \) and 0 otherwise. \( \mathcal{D}^0 \) is the set of negative examples, and \( \mathcal{D}^1 \) is the set of positive examples.

\section{Ablation study data structures}
\label{sec:iopo}
\paragraph{Input-only fine-tuning data instructions. } ~\texttt{"Is this text generated by LLM?"}

\paragraph{Prompt Only fine-tuning data instructions. } ~\texttt{"Prompt Inverter predicts prompt that could have generated the input texts. Is this prompt predicted by an input texts written by LLM?"}

\paragraph{Ablation Prompt. }~\texttt{"Text A is generated by an LLM. Determine whether Text B is also generated by an LLM using a similar prompt. Meanwhile, determine whether Text B could have been generated from Prompt C using an LLM. Answer with YES or NO."}

\section{DPIC (decouple prompt and intrinsic characteristics) Prompt Extraction Zero-shot Prompts~\cite{r62}}
\label{sec:DPIC prompt}
~\texttt{"I want you to play the role of the questioner. I will type an answer in English, and you will ask me a question based on the answer in the same language. Don’t write any explanations or other text, just give me the question. <TEXT>."}. 
\section{Comparison with DPIC}
\label{sec:compare dpic}
Since DPIC has not released its code, data, or models, we are unable to independently evaluate the performance of its classifier. Consequently, we rely on the reported results in the DPIC paper and construct a comparable dataset following their described settings to enable a fair comparison with IPAD. However, due to these limitations, we are unable to apply DPIC to additional datasets for broader evaluation.

To assess the generalization of IPAD, we reconstruct the following datasets, each containing 200 randomly sampled examples: \textbf{XSum}, \textbf{WritingPrompts}, and \textbf{PubMedQA}. For each dataset, we generate texts using three large language models: \texttt{ChatGPT (gpt-3.5-turbo)}, \texttt{GPT-4 (gpt-4)}, and \texttt{Claude 3 (claude-3-opus-20240229)}. Furthermore, the XSum datasets generated by these three models are augmented using two attack methods—\textbf{DIPPER} and \textbf{Back-Translation}—resulting in a total of 15 evaluation datasets.

\begin{table}[htbp]
\centering
\caption{AUROC comparison across tasks (XSum, Writing, PubMed) for ChatGPT, GPT-4, and Claude 3 using various prompt extraction methods.}
\label{tab:multi_task_accuracy}
\resizebox{\textwidth}{!}{
\begin{tabular}{l|cccc|cccc|cccc}
\hline
\textbf{Method} & \multicolumn{4}{c|}{\textbf{ChatGPT}} & \multicolumn{4}{c|}{\textbf{GPT-4}} & \multicolumn{4}{c}{\textbf{Claude 3}} \\
\cline{2-13}
 & XSum & Writing & PubMed & Avg. & XSum & Writing & PubMed & Avg. & XSum & Writing & PubMed & Avg. \\
\hline
DPIC (ChatGPT)     & 1.0000 & 0.9821 & 0.9092 & 0.9634 & 0.9996 & 0.9768 & 0.9438 & 0.9734 & 1.0000 & 0.9950 & 0.9686 & 0.9878 \\
DPIC (Vicuna-7B)   & 0.9976 & 0.9708 & 0.8990 & 0.9558 & 0.9986 & 0.9644 & 0.9394 & 0.9674 & 0.9992 & 0.9943 & 0.9690 & 0.9875 \\
IPAD (Version 1)   & 0.9850 & 0.9800 & 0.9250 & 0.9633 & 1.0000 & 0.9700 & 0.9700 & 0.9800 & 1.0000 & 0.9800 & 0.9750 & 0.9850 \\
IPAD (Version 2)   & 1.0000 & 0.9850 & 0.9800 & 0.9883 & 1.0000 & 0.9800 & 0.9500 & 0.9767 & 1.0000 & 0.9950 & 1.0000 & 1.0000 \\
\hline
\end{tabular}
}

\end{table}

\begin{table}[htbp]
\centering
\caption{AUROC comparison under generation perturbation settings (DIPPER, Back-translation) for each model.}
\label{tab:perturbation_accuracy}
\resizebox{\textwidth}{!}{
\begin{tabular}{l|ccc|ccc|ccc}
\hline
\textbf{Method} & \multicolumn{3}{c|}{\textbf{ChatGPT}} & \multicolumn{3}{c|}{\textbf{GPT-4}} & \multicolumn{3}{c}{\textbf{Claude 3}} \\
\cline{2-10}
 & Ori. & DIPPER & Back-trans. & Ori. & DIPPER & Back-trans. & Ori. & DIPPER & Back-trans. \\
\hline
DPIC (ChatGPT)     & 1.0000 & 1.0000 & 0.9972 & 0.9996 & 0.9991 & 0.9931 & 1.0000 & 0.9996 & 0.9878 \\
DPIC (Vicuna-7B)   & 0.9976 & 0.9980 & 0.9889 & 0.9986 & 0.9969 & 0.9903 & 0.9992 & 0.9996 & 0.9979 \\
IPAD (Version 1)   & 0.9850 & 0.8900 & 0.9850 & 1.0000 & 0.8950 & 0.9900 & 1.0000 & 0.9250 & 0.9950 \\
IPAD (Version 2)   & 1.0000 & 0.9750 & 0.9950 & 0.9800 & 0.9750 & 0.9950 & 1.0000 & 1.0000 & 1.0000 \\
\hline
\end{tabular}
}

\end{table}
Based on the experimental results, IPAD performs well and exhibits notable resistance to adversarial attacks.

IPAD open-sourced all the fine-tuned models, including the Prompt Inverter, and the two versions of distinguishers. Therefore, all the experiment results can be validated and reproduced.

\section{IPAD and DPIC prompt inverter examples}

\clearpage
\begin{center} 
\begin{longtable}{p{9cm} p{3cm} p{3cm}}
    \caption{IPAD and DPIC prompt inverter examples} \\
    \hline
    \textbf{Input} & \textbf{IPAD} & \textbf{DPIC} \\
    \hline
    \endfirsthead
    \hline
    \textbf{Input} & \textbf{IPAD} & \textbf{DPIC} \\
        \hline
        \endhead
{\small The IPC opened proceedings against the National Paralympic Committee of Russia after a report claimed the country had operated a widespread doping programme.A decision on any ban will come in the week commencing 1 August.The International Olympic Committee (IOC) has opted against a blanket ban."I can assure you that our board will take the right decision in the interest of sport and the interest of the Paralympic movement moving forward," said Craven.On Sunday, the IOC said it would leave it up to the governing bodies of individual sports to decide if Russian competitors are clean and should be allowed to take part.But Craven, himself a member of the IOC, was critical of that decision and said the IPC would not necessarily follow suit."I am disappointed in their decision, but that is a personal view," he added."We have to acknowledge their right to take such a decision. This is ultra-serious. I don't think there has been a situation in the past where you have had institutional doping on such a scale."We believe the Russian NPC is either unwilling or unable to uphold the IPC anti-doping code, which is in line with the World Anti-Doping Agency code, so that is what they have to respond to."Canadian law professor Richard McLaren's report, published last week, claimed Russia operated a state-sponsored doping programme from 2011 to 2015.The IPC said it acted after McLaren provided the names of the athletes associated with the 35 "disappearing positive samples" from the Moscow laboratory highlighted in the report.Nineteen samples potentially doctored as part of the sample-swapping regime during the 2014 Sochi Paralympic Winter Games have been sent for further analysis.Russia will have up to 21 days to appeal against any IPC decision, with the Rio Paralympics due to begin on 7 September.}&
{\small Explain the actions taken by the International Paralympic Committee against the Russian Paralympic Committee and the reasons behind these actions.} & {\small What were the key findings of Richard McLaren's report regarding Russia's doping programme?} \\
        \hline
{\small The world came crashing down in minutes. Many of us were asleep when it happened, and did n' t find out about it until later. When we awoke, we saw the carnage spread through the land, and we wept. There were n' t many of us left, but what few there were managed to find each other over the Internet. We gathered together in what remained of a major city on the East Coast of what was once the United States. It took us time, but we eventually began to rebuild. The brightest among those who survived thought to ask " Why, " while most of us were content with just surviving. Years passed, and no link was found between us. Eventually, those who had the question resigned themselves to the fact that they would never know. They went to their new homes, and tried to integrate themselves as best they could into the new society. It was n' t until 14 years after the event happened that the connection was discovered, quite by accident. One of the former questioners had taken a job as a mover, and was helping a fellow survivor move into a newly cleaned house. Sticking out of one of the boxes was the missing link. " Oh, you used to shop at Bad Dragon too? " Moments later, the realization struck him.}&{\small Write an essay about a time when you and your friends were the only survivors of a catastrophic event that wiped out most of the world's population. Describe how you and your friends coped with the aftermath and the challenges you faced in rebuilding society.}& {\small What was the event that caused the world to come crashing down?}\\
\hline
{\small In an alternate timeline, a second sentient race evolved in parallel with humans. These beings, known as the Avralians, possessed extraordinary abilities and resided in the hidden corners of the Earth. For centuries, unaware of each other's existence, humans and Avralians progressed independently. However, fate intervened one fateful day when an Avralian girl named Elara stumbled upon a group of human explorers deep in the forest. With wide-eyed wonder, she approached them cautiously. The humans, initially startled, soon recognized Elara's unique nature. They befriended her, and through her, the two races began to unravel the secrets of their intertwined existence. As time passed, harmony emerged between the two races, and mutual respect solidified their bond. The Avralians shared their ancient knowledge and wisdom, while humans brought innovation and technology. Together, they worked towards a world where unity triumphed over differences. However, not all were convinced of this newfound alliance. A faction within the human population, fueled by fear and mistrust, sought to exploit Avralian abilities for personal gain. Conflict brewed, threatening the fragile equilibrium. Elara, fueled by her pure heart and belief in unity, emerged as a beacon of hope. With a small band of supporters, she embarked on a journey to bridge the gap and foster understanding. Battles were fought, sacrifices made, but ultimately, Elara's message prevailed. Humans and Avralians learned to cherish their diversity and forge a future marked by collaboration and empathy. The world transformed into a tapestry of coexistence, where magnificent cities stood as testaments to unity and cultural exchange. Humans and Avralians moved freely through bustling markets, sharing knowledge, stories, and laughter. Together, they faced global challenges, from climate crises to epidemics, with unwavering determination.}&{\small Write an essay describing an alternate timeline in which a second sentient race evolved in parallel with humans, exploring the potential interactions and conflicts between the two species.}&{\small How did Elara manage to convince both races to embrace unity despite the conflict?}\\
\hline
{\small Both times I had the banana pepper appetizer, which is great and goes really well with the FRESH and delicious bread and cheese they give you at the start of your meal. nnFor entrees, me and my girlfriend have had mixed experience. I've had the fish sandwich (very good) and the eggplant parm sandwich (okay). My girlfriend got the salad with bread and basil on it, but the basil was over powering and the bread was soggy with the dressing. nnThe service is also a mixed bag. The first time our server went out of her way to take care of us and even MADE me cocktail sauce for my fish sandwich. The second time, the server was lackluster, didn't know anything about the menu and wasn't able to take proper care of us. nnI would return to Papa J's, but I my terrible experience last time isn't enough to say it would be my first pick of places to eat around Carnegie/Robinson.}&{\small This was a great place to stop for a quick lunch. The lines were not too long for the sandwiches they had and they had a wide selection of bagels if you wanted a bagel sandwich. With a great front patio for enjoying your food, it was a relaxing place to stop. Write a review for it.}&{\small What made the banana pepper appetizer stand out to you compared to other starters?}\\
\hline
{\small Abstract: This article explores the longstanding debate between Einstein's theory of general relativity and Maxwell's theory of electromagnetism regarding the nature of gravitation. The central question addressed is whether gravitation is best understood as a curvature of space, a field in flat space, or perhaps a combination of both concepts. Drawing upon a comprehensive analysis of the theoretical framework and empirical evidence, the article presents a nuanced examination of the arguments put forth by Einstein and Maxwell.The article begins by discussing Einstein's general theory of relativity, which proposes that gravitation arises from the curvature of spacetime caused by mass and energy. It outlines the mathematical formalism used to describe this curvature and highlights the key predictions and experimental confirmations of the theory. Conversely, the article delves into Maxwell's electromagnetic theory, which suggests that gravitation may be explained as a fundamental force mediated by a field propagating through flat space, similar to electromagnetic fields.Further, the article explores the distinctive features and limitations of each theory. It scrutinizes the conceptual foundations, mathematical rigor, and empirical support for both approaches, highlighting their respective strengths and weaknesses. Moreover, the article examines attempts to reconcile the two theories into a unified framework, such as the development of theories of quantum gravity.By critically evaluating the arguments and evidence from both camps, this article aims to offer a comprehensive assessment of the question regarding the nature of gravitation. Based on the analysis presented, it becomes evident that both Einstein's theory of general relativity and Maxwell's theory of electromagnetism provide valuable insights into the phenomenon of gravitation.}&{\small Write a paper abstract to explain the debate between Einstein's theory of general relativity and Maxwell's theory of electromagnetism regarding the nature of gravitation, and argue for which theory is more likely to be correct based on the evidence presented in the essay statement.}&{\small What are the main challenges in reconciling Einstein’s theory of general relativity with Maxwell’s theory of electromagnetism in explaining gravitation?}\\
\hline
    \end{longtable}
    \end{center}


\newpage
\section*{NeurIPS Paper Checklist}

\begin{enumerate}

\item {\bf Claims}
    \item[] Question: Do the main claims made in the abstract and introduction accurately reflect the paper's contributions and scope?
    \item[] Answer: \answerYes{} 
    \item[] Justification: We explained the contributions listed in the abstract and introduction in Section 2-4.

\item {\bf Limitations}
    \item[] Question: Does the paper discuss the limitations of the work performed by the authors?
    \item[] Answer: \answerYes{} 
    \item[] Justification: We present the limitation in the section Conclusion.

\item {\bf Theory assumptions and proofs}
    \item[] Question: For each theoretical result, does the paper provide the full set of assumptions and a complete (and correct) proof?
    \item[] Answer: \answerNA{} 
    \item[] Justification: There is no theorem or lemma in this paper.

    \item {\bf Experimental result reproducibility}
    \item[] Question: Does the paper fully disclose all the information needed to reproduce the main experimental results of the paper to the extent that it affects the main claims and/or conclusions of the paper (regardless of whether the code and data are provided or not)?
    \item[] Answer: \answerYes{}
    \item[] Justification: The codes are presented in the anonymous github, and the results can be reproduced by following the readme.

\item {\bf Open access to data and code}
    \item[] Question: Does the paper provide open access to the data and code, with sufficient instructions to faithfully reproduce the main experimental results, as described in supplemental material?
    \item[] Answer: \answerYes{}
    \item[] Justification: The codes are presented in the anonymous github, and the results can be reproduced by following the readme. The dataset for this paper are publicly available.

\item {\bf Experimental setting/details}
    \item[] Question: Does the paper specify all the training and test details (e.g., data splits, hyperparameters, how they were chosen, type of optimizer, etc.) necessary to understand the results?
    \item[] Answer: \answerYes{} 
    \item[] Justification: The experimental details are presented.

\item {\bf Experiment statistical significance}
    \item[] Question: Does the paper report error bars suitably and correctly defined or other appropriate information about the statistical significance of the experiments?
    \item[] Answer: \answerNo{} 
    \item[] Justification: We set the temperature parameter of the large language models to 0, so the results are deterministic.

\item {\bf Experiments compute resources}
    \item[] Question: For each experiment, does the paper provide sufficient information on the computer resources (type of compute workers, memory, time of execution) needed to reproduce the experiments?
    \item[] Answer: \answerYes{} 
    \item[] Justification: We use one A800-80GB gpu and one Nvidia v100 to conduct our experiments.
    \begin{itemize}
        \item The answer NA means that the paper does not include experiments.
        \item The paper should indicate the type of compute workers CPU or GPU, internal cluster, or cloud provider, including relevant memory and storage.
        \item The paper should provide the amount of compute required for each of the individual experimental runs as well as estimate the total compute. 
        \item The paper should disclose whether the full research project required more compute than the experiments reported in the paper (e.g., preliminary or failed experiments that didn't make it into the paper). 
    \end{itemize}
    
\item {\bf Code of ethics}
    \item[] Question: Does the research conducted in the paper conform, in every respect, with the NeurIPS Code of Ethics \url{https://neurips.cc/public/EthicsGuidelines}?
    \item[] Answer: \answerYes{} 
    \item[] Justification: This paper has no ethics problem.

\item {\bf Broader impacts}
    \item[] Question: Does the paper discuss both potential positive societal impacts and negative societal impacts of the work performed?
    \item[] Answer: \answerYes{} 
    \item[] Justification: We discuss the potential societal impacts mainly in section 3.2.
    
\item {\bf Safeguards}
    \item[] Question: Does the paper describe safeguards that have been put in place for responsible release of data or models that have a high risk for misuse (e.g., pretrained language models, image generators, or scraped datasets)?
    \item[] Answer: \answerNA{} 
    \item[] Justification: There are no such risks in our paper.
    
\item {\bf Licenses for existing assets}
    \item[] Question: Are the creators or original owners of assets (e.g., code, data, models), used in the paper, properly credited and are the license and terms of use explicitly mentioned and properly respected?
    \item[] Answer: \answerYes{} 
    \item[] Justification: : We have cited the original papers and included proper license.

\item {\bf New assets}
    \item[] Question: Are new assets introduced in the paper well documented and is the documentation provided alongside the assets?
    \item[] Answer: \answerNA{} 
    \item[] Justification: We do not release new assets.

\item {\bf Crowdsourcing and research with human subjects}
    \item[] Question: For crowdsourcing experiments and research with human subjects, does the paper include the full text of instructions given to participants and screenshots, if applicable, as well as details about compensation (if any)? 
    \item[] Answer: \answerNA{} 
    \item[] Justification: We have no crowdsourcing experiments and human subjects.

\item {\bf Institutional review board (IRB) approvals or equivalent for research with human subjects}
    \item[] Question: Does the paper describe potential risks incurred by study participants, whether such risks were disclosed to the subjects, and whether Institutional Review Board (IRB) approvals (or an equivalent approval/review based on the requirements of your country or institution) were obtained?
    \item[] Answer: \answerNA{} 
    \item[] Justification: We have no crowdsourcing experiments and human subjects.

\item {\bf Declaration of LLM usage}
    \item[] Question: Does the paper describe the usage of LLMs if it is an important, original, or non-standard component of the core methods in this research? Note that if the LLM is used only for writing, editing, or formatting purposes and does not impact the core methodology, scientific rigorousness, or originality of the research, declaration is not required.
    \item[] Answer: \answerYes{} 
    \item[] Justification: Pretrained LLMs are used as backbones in our method, which is clearly stated in this paper.

\end{enumerate}

\end{document}